\def\year{2022}\relax
\documentclass[letterpaper]{article} 
\usepackage{aaai22}  
\usepackage{times}  
\usepackage{helvet}  
\usepackage{courier}  
\usepackage[hyphens]{url}  
\usepackage{graphicx} 
\usepackage{natbib}  
\usepackage{caption} 
\DeclareCaptionStyle{ruled}{labelfont=normalfont,labelsep=colon,strut=off} 
\usepackage{algorithm}
\usepackage{algorithmic}

\usepackage{newfloat}
\usepackage{listings}
\floatstyle{ruled}
\newfloat{listing}{tb}{lst}{}
\floatname{listing}{Listing}
\usepackage{amsmath,amsfonts,amssymb}
\usepackage{graphicx}
\usepackage{textcomp}
\usepackage{gensymb}
\usepackage{enumitem}
\usepackage{comment}
\usepackage{multirow}
\usepackage{subcaption}
\usepackage{booktabs}
\usepackage[switch]{lineno}

\newcommand{\dataname}{\textsc{Caise}}

\title{\dataname{}: Conversational Agent for Image Search and Editing}
\author {
Hyounghun Kim\textsuperscript{1}, 
Doo Soon Kim\textsuperscript{2},
Seunghyun Yoon\textsuperscript{3}\\
Franck Dernoncourt\textsuperscript{3},
Trung Bui\textsuperscript{3},
Mohit Bansal\textsuperscript{1}\\}
\affiliations{
\textsuperscript{1}UNC Chapel Hill \;\;\; 
\textsuperscript{2}Roku Inc. \;\;
\textsuperscript{3}Adobe Research\\
\{hyounghk, mbansal\}@cs.unc.edu\\
\{syoon, dernonco, bui\}@adobe.com\\
}
\usepackage{bibentry}

\begin{document}
\maketitle

\begin{abstract}
  Demand for image editing has been increasing as users' desire for expression is also increasing. However, for most users, image editing tools are not easy to use since the tools require certain expertise in photo effects and have complex interfaces. Hence, users might need someone to help edit their images, but having a personal dedicated human assistant for every user is impossible to scale. For that reason, an automated assistant system for image editing is desirable. Additionally, users want more image sources for diverse image editing works, and integrating an image search functionality into the editing tool is a potential remedy for this demand. Thus, we propose a dataset of an automated \textbf{C}onversational \textbf{A}gent for \textbf{I}mage \textbf{S}earch and \textbf{E}diting (\dataname{}). To our knowledge, this is the first dataset that provides conversational image search and editing annotations, where the agent holds a grounded conversation with users and helps them to search and edit images according to their requests. To build such a system, we first collect image search and editing conversations between pairs of annotators. The assistant-annotators are equipped with a customized image search and editing tool to address the requests from the user-annotators. The functions that the assistant-annotators conduct with the tool are recorded as executable commands, allowing the trained system to be useful for real-world application execution. We also introduce a generator-extractor baseline model for this task, which can adaptively select the source of the next token (i.e., from the vocabulary or from textual/visual contexts) for the executable command. This serves as a strong starting point while still leaving a large human-machine performance gap for useful future work.\footnote{Our code and dataset are publicly available at: https://github.com/hyounghk/CAISE}
\end{abstract}

\begin{figure}[t]
    \centering
    \includegraphics[width=0.95\columnwidth]{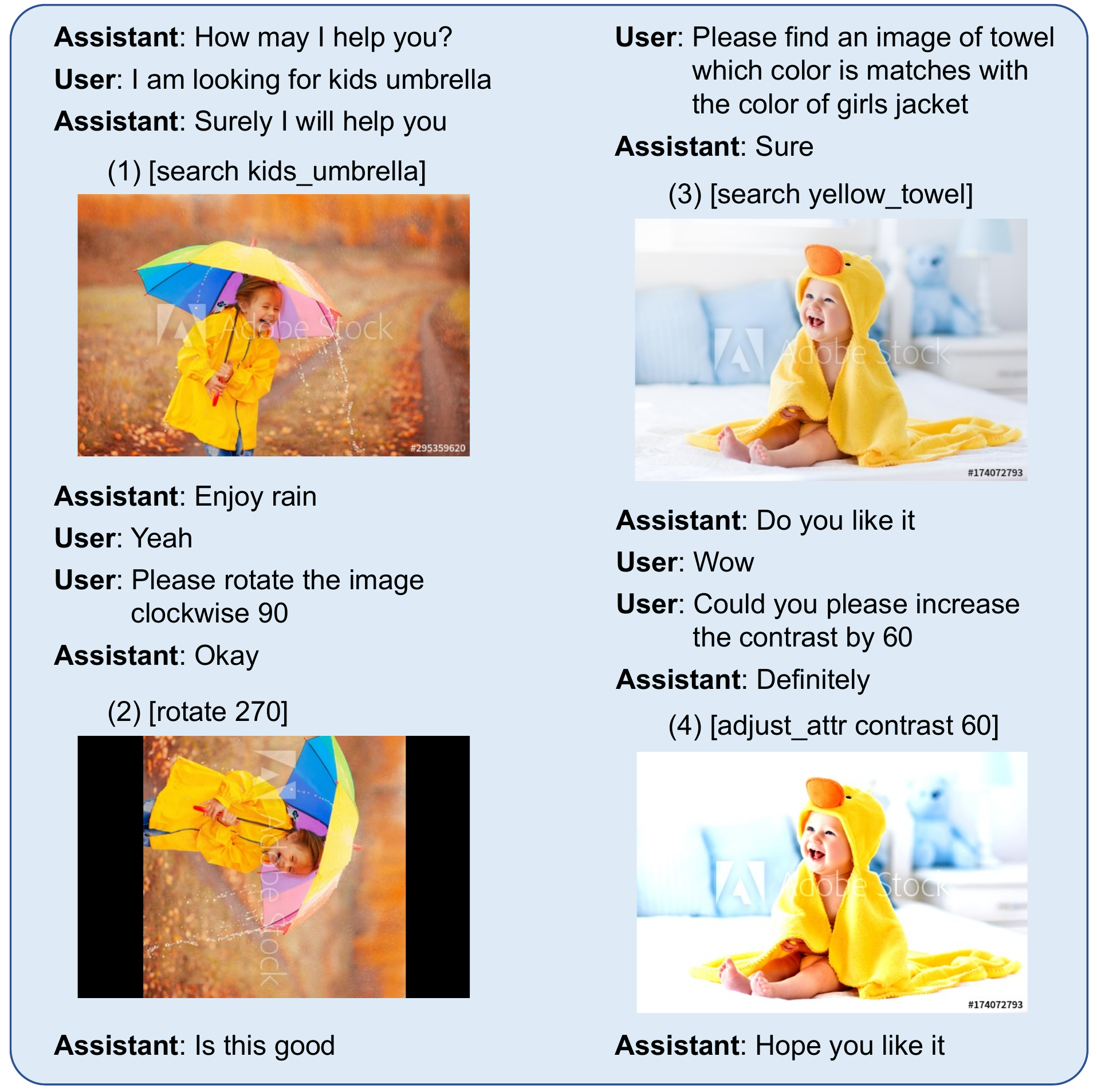}
    \caption{Conversational agent for image search and editing (\dataname{}). The dialogue starts with the image search request from the user. The assistant conducts the image search and addresses the following image search or editing requests for the user through 4 turns of request-execution exchange ([$\cdot$] shows the image search/editing commands to the system).
    }
    \label{fig:dialog_fig}
\end{figure}

\begin{figure*}[t]
    \centering
    \includegraphics[width=1.95\columnwidth]{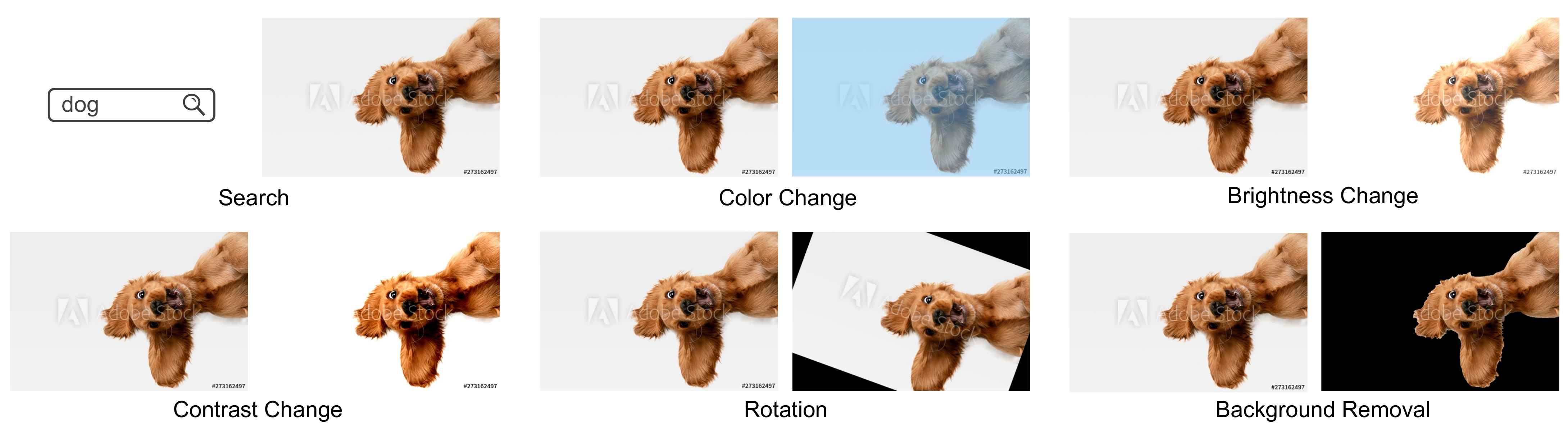}
    \caption{The diverse image search and editing effect functions that our \dataname{} dataset employs.
    \label{fig:editing_effects}}
\end{figure*}

\section{Introduction} 
As the technology of image editing is developing and being refined, its utility is also increasing. It has become a usual practice to add editing effects to photos to make them look better. However, using image editing tools requires the expertise and skill that regular layperson users do not have. The names of these photo effects are not familiar and even the implication of the effects on images are not intuitive for most users. Hence, to increase the accessibility to these tools, proper individual expert guidance is required. However, guidance assistant systems run by small groups of available human experts could not cover all the requests from a large number of users worldwide. Instead, editing tools can benefit from having an automated assistant system that can have a conversation with users at scale to help them with their editing needs. 

On the other hand, as the purpose and use cases of image editing are getting diverse, source materials for image editing also need to be diversified. For example, users might want to recreate their photos by adding additional objects from external sources. Users may also want to follow a reference image to make their photos more attractive (e.g., by borrowing a color from the source image). Hence, these activities call for an image search interface to be integrated with image editing tools to provide a more integrated and comprehensive platform.

There have been some prior efforts towards automated image editing systems. They have focused on intent/action/goal
identification from image editing requests~\cite{manuvinakurike2018dialedit, manuvinakurike2018edit,manuvinakurike2018conversational,lin2020multimodal}, exploring low-level editing terms~\cite{lin2020adjusting}, editing images from descriptions~\cite{shi2020benchmark}, and describing image differences caused by image editing~\cite{tan2019expressing}. However, there has been limited effort to integrate conversational image search and editing functions in a directly executable end-to-end manner for deployment into real-world applications. Therefore, we propose a new dataset, \dataname{} (`\textbf{C}onversational \textbf{A}gent for \textbf{I}mage \textbf{S}earch and \textbf{E}diting'), in which a user and an assistant hold a conversation in natural language (English) about image search and editing (Figure~\ref{fig:dialog_fig}). The user's role is to make requests for image search and editing and the assistant's role is to search or edit images according to the user's requests and return the results while responding with natural language.

To collect such data, we implement a dialogue interface and ask pairs of annotators (one operating as the user and the other one as the assistant) to converse and search/edit images via the interface. The user is provided with multiple seed images from which they can get some ideas about what to search in the first place. Also, we show the user a list of suggested image search/editing functions to keep the command types diverse by asking them to follow the list as long as they can. The assistant annotator, on the other hand, is equipped with an image search and editing interface to perform the user's requests. All command executions lead to the corresponding executable commands to be recorded. A total of 1.6K dialogues and 6.2K task instances are collected. The collected dialogues contain different types of image search/editing requests from users (direct request, implied request, object referring request; Table~\ref{tbl:req_examples}) and assistants' diverse responses (Sec.\ref{sec:assis_resp}), requiring models to understand the diverse grounded interactions in the conversations.

The task on the \dataname{} dataset is to generate the executable commands (e.g., search, color-change, brightness-change, contrast-change, rotation, background-removal; Figure~\ref{fig:editing_effects}) given the conversation and image contexts. This task setup simulates real-world image editing tools, facilitating important initial steps towards deployment in downstream applications. We introduce a novel generator-extractor model as a strong starting point baseline for this task and dataset. We employ a copying mechanism~\cite{vinyals2015pointer, gu2016incorporating, miao2016language, see2017get}, with which the model adaptively selects a way (i.e., generate from the vocabulary or extract from the context) to decode the next word since the clues for arguments of an executable command could be implicitly mentioned in the user's request (e.g., \emph{``Please change the image color with \textbf{color of bus}''}) or the request contains the direct cues (e.g., \emph{``Is it possible to increase the brightness of the image by \textbf{30} percent''}). For more effective model performance, we extend this mechanism so that it can also cover visual concepts in images by extracting object attributes or names from a set of object detection based concepts. For example, for the request \emph{``Please change the image color with the color of bus''}, the corresponding color can not only be generated from the vocabulary, but also directly copied from one of object detection results, \emph{``red bus''}. Our experiments show our baseline model performs effectively as a starting point, and we also demonstrate a large human-machine performance gap to allow useful future works on this important and understudied task.

Our contributions are two-fold: (1) we introduce a novel grounded dialogue dataset, \dataname{}, which incorporates image search and editing, featuring executable commands, hence allowing for more practical use in real-world applications. (2) We also introduce a generator-extractor model as a strong starting point baseline which extends the copy mechanism to the visual concept extraction, allowing for more effective performance and helping the interpretation of the model's behavior, while also leaving a large human-machine performance gap to allow useful future work by the community on this new challenging multimodal task.

\section{Related Work}

\vspace{3pt}
\par
\noindent\textbf{Image Editing.}
There have been some prior efforts to automate image editing programs. The research on image editing has been focused on intent identification~\cite{manuvinakurike2018conversational}, request to actionable command mapping~\cite{manuvinakurike2018edit, lin2020multimodal}, dialogue act labeling~\cite{manuvinakurike2018dialedit}, low-level image edit requests~\cite{lin2020adjusting}, description to editing~\cite{shi2020benchmark}, or editing to description~\cite{tan2019expressing}. Also, language-based image editing~\cite{shinagawa2018interactive, chen2018language, el2019tell, fu2020sscr} focuses on an image generation task setup. However, there have been relatively few studies that pursue end-to-end conversational image editing agent systems combined with image search functionality. Our \dataname{} dataset supports the direct deployment of conversational image editing assistant systems by incorporating executable commands in the dataset, and also integrates image search functionality so as to make it more comprehensively useful.

\vspace{3pt}
\par
\noindent\textbf{Referring Expression Comprehension.}
Referring to an object using neighboring objects and relations between them is important to specify the object exactly and reduce ambiguities. Agents should have the ability to understand these expressions for better communication with humans or other agents. Referring expression comprehension has been studied actively~\cite{kazemzadeh2014referitgame, mao2016generation, hu2016natural, yu2018mattnet, Chen19:touchdown, qi2019rerere}. Object referring expression plays an important role in image search and editing activities too. Users might need to specify an object or region that photo effects should be applied to, or want to search an item, of which they don't know the exact name, by referring it using spatial relations with other objects in an image. Our \dataname{} dataset contains a large amount of referring expressions to encourage agents to have the ability to understand such expressions.

\vspace{3pt}
\par
\noindent\textbf{Multimodal Dialogue.}
Multimodal dialogue has been actively studied in previous works~\cite{visdial, de2017guesswhat, mostafazadeh2017image, saha2017towards, pasunuru2018game, alamri2019audio, haber2019photobook, kim2019codraw, moon2020situated, shuster2020image, cheng2020sequential}. Although all these works involve interesting task setups (question answering/generation, object discovery, shopping, collaborative drawing, response retrieval/generation, image identification/generation, etc.) with different multimodal features (text, image, video, audio), there has not been a focus on how to generate directly executable commands from the grounded multimodal dialogue. Moreover, to the best of our knowledge, our \dataname{} dataset and task is the first large-scale multimodal dialogue setup which combines the image search and editing tasks.

\section{Task Description}
Multimodal dialogue based executable command generation is one task that can be introduced from our \dataname{} dataset. Specifically, given a conversation history, previously searched and edited images, and previously executed commands, the agent should generate an executable command which can return the correct result for the user's request. The definitions of the executable commands are as follows:

\vspace{3pt}
\par
\noindent\textbf{Search.}
The search command retrieves images that are searched online with a query string. The format of the search command is [\emph{search argument\_1 ... argument\_n ...}]. \emph{`argument\_n'} is the n-th token in the query string and there is no limit for the number of arguments.

\vspace{3pt}
\par
\noindent\textbf{Color Change.}
The color change command paints a whole image with a designated color. The format of the color change command is [\emph{adjust\_color argument\_1 argument\_2}]. \emph{`argument\_1'} is a name of the colors (red, orange, green, blue, sky blue, purple, brown, yellow, pink), and \emph{`argument\_2'} is the value of intensity (0.0-1.0).

\vspace{3pt}
\par
\noindent\textbf{Brightness Change.}
The brightness change command changes the brightness of a whole image with a designated intensity. The format of the brightness change command is [\emph{adjust\_attr brightness argument\_1}]. \emph{`argument\_1'} is the value of intensity (-100-100\%).

\vspace{3pt}
\par
\noindent\textbf{Contrast Change.}
The contrast change command changes the contrast of a whole image with a designated intensity. The format of the contrast change command is [\emph{adjust\_attr contrast argument\_1}]. \emph{`argument\_1'} is the value of intensity (0-100\%).

\vspace{3pt}
\par
\noindent\textbf{Rotation.}
The rotation command rotates a whole image by a designated degree. The format of the rotation command is [\emph{rotate argument\_1}]. \emph{`argument\_1'} is the value of degree (0-360).

\vspace{3pt}
\par
\noindent\textbf{Background Removal.}
The background removal command makes a whole image black except the main subject. The format of the background removal command is [\emph{image\_cutout}]. There is no argument.
\\
\\
For illustrations of these photo effects, see Figure~\ref{fig:editing_effects}.

\section{Dataset}
Our \dataname{} dataset consists of conversations between a `user' and an `assistant'. Each conversation includes utterances of the user and assistant, searched or edited images, and executed commands.

\vspace{3pt}
\par
\noindent\textbf{Conversation Interface.}
We implement a dialogue system through which a pair of people chat about image search and editing. We build the user-side and the assistant-side interfaces separately since their roles are quite different. In the user-side interface, we provide 15 random seed images from COCO dataset~\cite{lin2014microsoft} to help the user decide what to request for the first image search. We also present a suggestion for types of search and editing, which is a list of four commands from different types being randomly selected and ordered to avoid repeating the same search/editing order so that the user can follow it when they request to the assistant. In the assistant-side interface, we prepare a customized light-weight search and editing tool for the assistant to address the users' requests. We use Adobe Stock\footnote{ \url{https://www.adobe.io/apis/creativecloud/stock.html} (the watermarks on the images are from using the Adobe Stock API).} for the image search engine, Adobe Photoshop\footnote{ \url{https://adobedocs.github.io/photoshop-api-docs-pre-release/}} for the background removal function, and OpenCV\footnote{\url{https://opencv.org/}} to implement the other editing functions. All the search and editing effects conducted from the tool are recorded in the form of executable commands that are used for corresponding functions. See Appendix~\ref{app:interface} for the images of the interfaces.

\begin{table}[t]
\begin{center}
\resizebox{0.85\columnwidth}{!}{
 \small
 \begin{tabular}{lcc}
 \toprule
  \multirow{2}{*}{}& \multicolumn{2}{c}{Count}\\
 \cmidrule{2-3}
  & Per Dialogue & Total\\
 \midrule
    Dialogue & - & 1,611 \\
  \midrule
  Utterance & 15.5 & 24,938 \\
  \midrule
  Utterance (user) & 7.9 & 12,641 \\
  \midrule
  Utterance (assistant) & 7.6 & 12,297\\
  \midrule
  Executable Command & 3.8 & 6,173 \\
  \midrule
  Image & 3.8 & 6,173 \\
  \bottomrule
\end{tabular}
}
\end{center}
\caption{The number of dialogue components. Dialogues in our \dataname{} dataset are long (15.5 utterances) with four turns of image search/editing request-execution exchanges. \label{tbl:dialog_stat}} 
\end{table}

\begin{table}[t]
\begin{center}
\resizebox{0.95\columnwidth}{!}{
 \small
 \begin{tabular}{lccccc}
 \toprule
  \multirow{2}{*}{}& \multicolumn{5}{c}{Length}\\
 \cmidrule{2-6} 
  & avg & stddev & median & max & min\\
 \midrule
  Utterance & 5.26 & 4.98 & 4.0 & 38  & 1 \\
  \midrule
  Utterance (user) & 6.99 & 6.16 & 6.0 & 38  & 1\\
  \midrule
  Utterance (assistant) & 3.49 & 2.24 & 3.0 & 24 & 1\\
  \bottomrule
\end{tabular}
}
\end{center}
\caption{The lengths of utterances in the dialogue collection. The user utterances are longer than assistant's due to the difference in their roles. The standard deviation of the lengths is large, indicating the utterances have various lengths. \label{tbl:utter_length}} 
\end{table}

\vspace{3pt}
\par
\noindent\textbf{Data Collection.}
We employ 10 annotators and train them to make them familiar with the collection interfaces and their primary roles, and guarantee the quality of the dataset. In the training session, we check all the practice dialogues manually and give feedback. We perform this training session multiple times until the quality of the dialogues gets above some threshold (see Appendix~\ref{app:annotator} for the detailed training process). After the training period, two annotators are paired so that one of them takes the user role and the other takes the assistant role. User-annotators are asked to give four requests throughout a conversation. Assistant-annotators are asked to perform the image search and editing functions according to the user-annotators' requests. If the user-annotators' requests are not clear, the assistant-annotators can ask them to clarify. We hire freelancers since the collection process needs some training to build expertise (especially for manipulating the search/editing interface), and pairing between the user and assistant annotators via a general crowd-sourcing platform is not easy.\footnote{We use Upwork (\url{https://www.upwork.com}) to hire freelancer annotators for high-quality, trained-expert human feedback. Upwork provides various communication tools (text chat and video/audio call interfaces) to facilitate communication with annotators and thus enable effective and efficient annotator training, as also shown in \cite{stiennon2020learning}.}

\vspace{3pt}
\par
\noindent\textbf{Payment.}
We pay up to 2 USD per dialogue, including bonuses. We also pay for dialogues which are created by annotators in their training period. Considering the time taken for a dialogue (around 5 minutes for a pair of trained annotators), the hourly wage is competitive (nearly 12 USD/hour per annotator). 

\begin{table}[t]
\begin{center}
\resizebox{0.95\columnwidth}{!}{
 \begin{tabular}{cc}
 \toprule
  Type & Examples\\
 \midrule
 \multirow{4}{*}{Dir-Req} & \multirow{1}{*}{\shortstack[c]{\emph{``I was looking for an image of zoo''}}}\\
  & \multirow{2}{*}{\shortstack[c]{\emph{``Now increase the brightness} \\\emph{of the image by 40 percent''}}}\\
  & \\
  & \multirow{1}{*}{\shortstack[c]{\emph{``Please get rid of the background''}}}\\
    \midrule
  \multirow{2}{*}{Impl-Req} & \multirow{1}{*}{\shortstack[c]{\emph{``Can we repeat further by 130 degree more''}}}\\
  & \multirow{1}{*}{\shortstack[c]{\emph{``Can we try increasing further by 50 more''}}}\\
  \midrule
  \multirow{4}{*}{ObjRef-Req} & \multirow{2}{*}{\shortstack[c]{\emph{``Please find an image of the object seen} \\\emph{to the right of the juicer in the above image''}}}\\
  &\\
  & \multirow{2}{*}{\shortstack[c]{\emph{``Please change the color of image} \\\emph{which matches with the color of cushion''}}}\\
  &\\
  \bottomrule
\end{tabular}
}
\end{center}

\caption{The examples of different types of requests (Dir-Req: direct request, Impl-Req: implied request, ObjRef-Req: object referring request). \label{tbl:req_examples}} 
\end{table}

\begin{figure}[t]
    \centering
    \includegraphics[width=0.7\columnwidth]{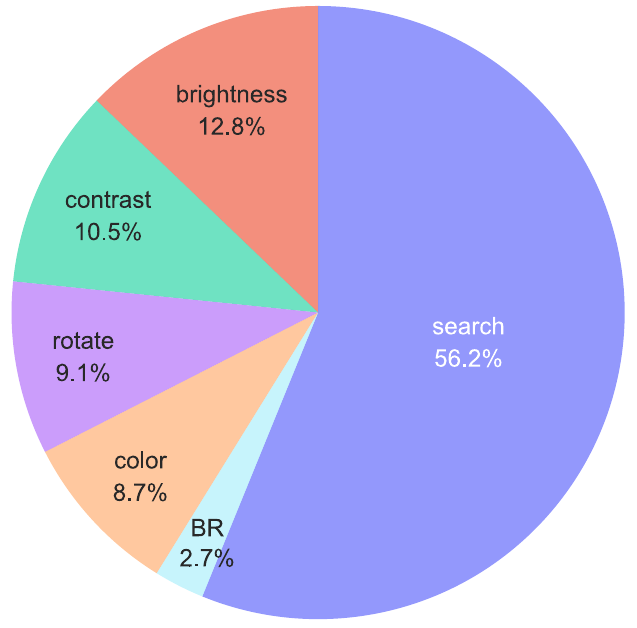}
    \caption{The executable commands frequency. The search command has the highest frequency since each dialogue begins with a search request (BR: background removal). \label{fig:api_pie}}
\end{figure}

\begin{figure*}[t]
    \centering
    \includegraphics[width=1.99\columnwidth]{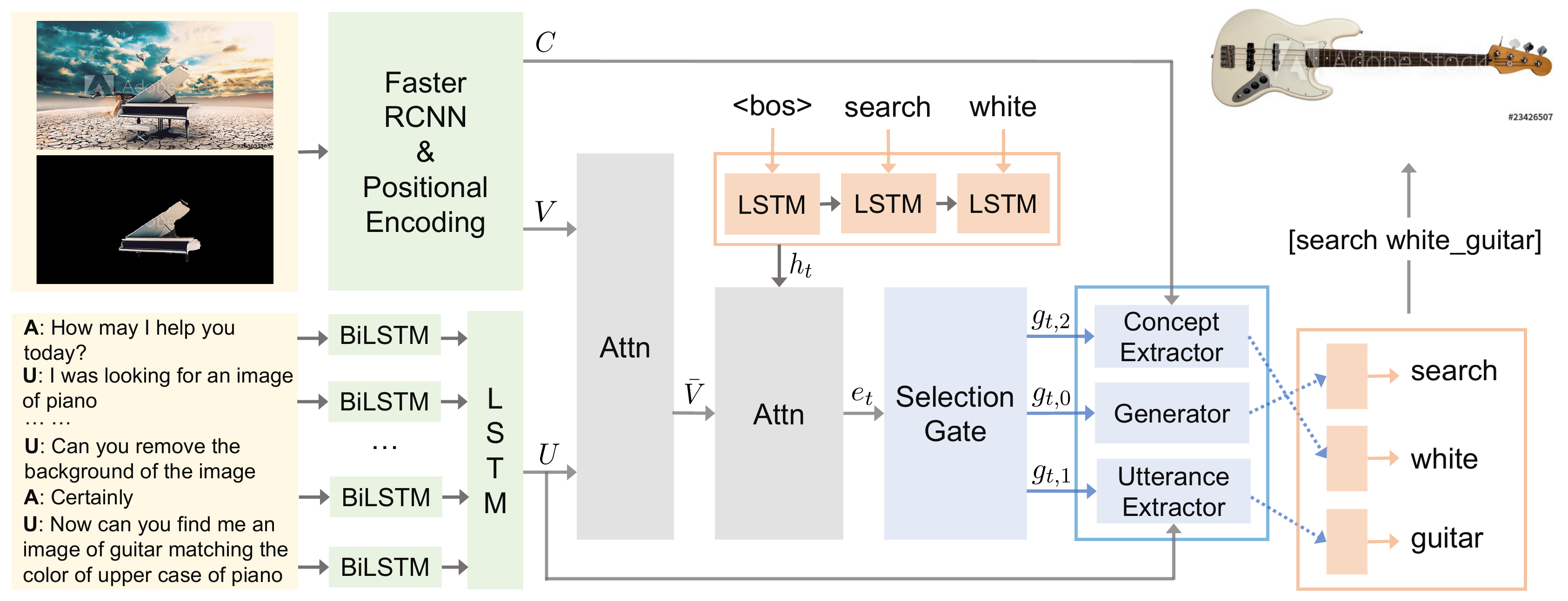}
    \caption{The Generator-Extractor model. The model adaptively selects the source of the next token via the selection gate (for the simplicity purpose, some blocks and relationship arrows are omitted; input ``$<$bos$>$ search white'' to the LSTM block is previously generated tokens, i.e., an autoregressive decoding setup; the model produces the command word-by-word). 
    }
    \label{fig:model_fig}
\end{figure*}

\section{Data Analysis}
We collect 1,611 dialogues and create 6,173 task instances from the dialogue collection (since each dialogue has around four executable commands). 

\vspace{3pt}
\par
\noindent\textbf{Dialogue Length.}
As shown in Table~\ref{tbl:dialog_stat}, the average number of utterances from both the users and assistants is 15.5 (7.9 and 7.6 from users and assistants, respectively). The average number of executable commands and images are the same (3.8 per dialogue) since each image is the result of the execution of each corresponding command.

\vspace{3pt}
\par
\noindent\textbf{Utterance Length.}
As shown in Table~\ref{tbl:utter_length}, the average length of user utterances is larger than that of assistant utterances (6.99 vs. 3.49). The reason is that user utterances are mainly about image search and editing requests, requiring detailed explanations (e.g., \emph{``Could you also find me an image of dress for my wife matching the color of hat in the above image?''}). On the other hand, assistant utterances are usually short responses to users' requests (e.g., \emph{``okay''}, \emph{``sure''}) or questions for users' confirmation (e.g., \emph{``Do you like it?''}, \emph{``Is this fine?''}), and clarifications (e.g., \emph{``clock wise or anti clockwise?''}). Also, the standard deviations of the utterance lengths are large compared to the average lengths, confirming utterances in our \dataname{} dataset have various lengths.

\vspace{3pt}
\par
\noindent\textbf{User Request Types.}
As shown in Table~\ref{tbl:req_examples}, we can categorize the image search and editing requests mainly into three types: direct request, implied request, and object referring request. Direct requests are self-contained requests which have direct clues about what users are asking. Implied requests are the ones that do not explicitly mention what types of functions are asked to be performed but imply them from the conversation contexts. Object referring requests are the ones that use the information of objects (i.e., color, name, location) in images to specify what should be done.

\vspace{3pt}
\par
\noindent\textbf{Assistant Response Types.\label{sec:assis_resp}}
Although several assistant responses are generic confirmation-based (since the assistants' main role is to perform image search and editing according to users' requests), there are also several other types of interesting responses such as correction (user: \emph{``... image of sea-saw ...''} - assistant: \emph{``Do you mean see-saw?''}), ambiguity-clarification (user: \emph{``... rotate the image to  40 degree''} -  assistant: \emph{``... clock wise or anti clockwise?''}), coreference (assistant: \emph{``How would you like it by''}), etc., encouraging models to understand the diverse grounded interactions in the conversations between users and assistants to perform the task.

\vspace{3pt}
\par
\noindent\textbf{Executable Commands Frequency.}
As shown in Figure~\ref{fig:api_pie}, the search command is the most frequent. The reason is that search is the first command that must be performed in every dialogue, and we design the collection interface so that each dialogue has one additional search command on average (the ratio of the first-line search commands to the other search commands is 46.5\% vs. 53.5\%). The low frequency of the background removal command (``BR'' in the figure) is due to the command's instability. Unlike the other commands that do not fail, the background removal command could fail depending on images' contents (it seems that images that have complicated contents are hard to remove background from). Once the background removal command fails, the user-annotators might not try it again and perform one of the other functions instead.

\begin{table*}[t]
\begin{center}
\resizebox{1.7\columnwidth}{!}{
 \begin{tabular}{clccccccc}
 \toprule
  &\multirow{2}{*}{Models} & \multicolumn{7}{c}{Accuracy (\%)}  \\
  \cmidrule{3-9}
  && total & search & color & brightness & contrast & rotation & remove-back \\
 \midrule
    1 & Base & 22.33 & 11.45 & 28.63 & 32.13 & 46.46 & 9.52 & \textbf{100.0}\\
    \midrule
     2 &Base+VE & 22.12 & 11.00 & 30.77 & 30.12 & 48.56 & 8.93 & \textbf{100.0} \\
    \midrule
    3 &Base+UE & 45.23 & 36.42 & 26.07 & \textbf{49.80} & 92.13 & \textbf{29.17} & 97.14 \\
    \midrule
     4 &Base+UE+VE & \textbf{46.43} & \textbf{37.43} & \textbf{40.60} & 48.39 & \textbf{93.18} & 26.49 & 97.14\\
    \midrule
    \midrule
    5 & Human Expert & 90.0 & 82.0 & 90.0 & 100.0 &100.0 & 100.0 & 100.0\\
    \bottomrule
\end{tabular}
}
\end{center}
\caption{Model performance on the test split. The extractors help improve the model's performance (Base: the basic encoder-decoder model only with generator (without extractors), UE: utterance extractor, VE: visual concept extractor).\label{tbl:result}} 
\end{table*}
\section{Models}
We present the generator-extractor model as a starting point baseline (Figure~\ref{fig:model_fig}). The model takes the history of utterances, images, and previously executed commands as input, and predicts a next executable command.

\vspace{3pt}
\par
\noindent\textbf{Encoder.}
A large part of our \dataname{} dataset involves objects and their concepts (names and attributes) in images, especially for the search command. Thus, we employ Faster R-CNN~\cite{girshick2015fast} to extract object visual features $\hat{V}$, bounding box features $B$, and their concept features $W^c$, which are usually made of a couple of tokens, from images $I$. $\hat{V}$ and $B$ are combined through a linear layer, and $W^c$ is further encoded by a word embedding layer and the bidirectional LSTM~\cite{hochreiter1997long}:
\begin{align}
    \hat{V}, B, W^c &= \textrm{FRCNN}(I),\;\;\;
    V = \textrm{PE}(\textrm{Linear}([\hat{V};B]))\\
    \hat{C} &= \textrm{Embed}(W^c), \;\;
    C = \textrm{PE}(\textrm{BiLSTM}(\hat{C}))
\end{align}
where PE denotes positional encoding~\cite{gehring2017convolutional, vaswani2017attention} and it is applied image-wise (i.e., the same encoding value is applied to the features from the same image). Sequences of tokens from utterances $W^u$ in dialogue $D$ are encoded by the bidirectional LSTM, and the last forward hidden state and the first backward hidden state of $\hat{U} \in \mathbb{R}^{M \times N \times d}$ are extracted and concatenated to create a vector which represents each utterance, where $M$ is the dialogue length, $N$ is the utterance length, and $d$ is the feature dimension. Then, the sequence of the utterance features is fed to a LSTM to learn the dialogue context:
\begin{align}
    \hat{U} &= \textrm{BiLSTM}(\textrm{Embed}(W^u)) \\ 
    U &= \textrm{LSTM}([\hat{U}_{N-1}^f;\hat{U}_0^b])
\end{align}
We employ the attention mechanism to align the visual features $V$, and utterance features $U\in \mathbb{R}^{M \times d}$.
We calculate the similarity matrix $S \in \mathbb{R}^{O \times M}$ between visual and utterance features, where $O$ is the total number of all object features from the images: $S_{ij} = V_i^{\top}U_j$. From the similarity matrix, the new fused visual and utterance feature is:
\begin{align}
    \bar{U} &= \textrm{softmax}(S) \cdot U, \;\;\;
    \bar{V} = [V;\bar{U};V \odot \bar{U}] \cdot W_v
\end{align}
where $W_v \in \mathbb{R}^{3d \times d}$ is the trainable parameter, $\odot$ is element-wise product, and $\cdot$ is matrix multiplication.
Tokens from an executable command, \(\{w_t\}_{t=1}^T\), are embedded in the embedding layer, and then sequentially fed to the LSTM layer.
\begin{align}
    \hat{w}_{t-1} &= \textrm{Embed}(w_{t-1}), \;\;\;
    h_{t} = \textrm{LSTM}(\hat{w}_{t-1}, h_{t-1})
\end{align}
The same (but with different parameters) attention mechanism (Attn), which is applied to visual and utterance features, is used for aligning the command feature, $h_{t}$, and $\bar{V}$.
\begin{align}
    e_t &= \textrm{Attn}(h_{t}, \bar{V})
\end{align}

\vspace{3pt}
\par
\noindent\textbf{Generator.}
The generator calculates the probability of each token in the vocabulary that contains all possible candidates.
\begin{align}
    l_t = \textrm{Linear}(e_t),\;\;\; a^g_t = \textrm{softmax}(l_t)
\end{align}

\vspace{3pt}
\par
\noindent\textbf{Extractor.}
Utterances in our \dataname{} dataset contain many direct clues for generating commands. Thus, the model would benefit from extracting keywords from the context. We employ a copying mechanism~\cite{vinyals2015pointer, gu2016incorporating, miao2016language, see2017get} to implement the extraction. 
\begin{align}
    (A^u_t)_{i} = h_t^{\top} U_i, \;\;\;
    a^u_t = \textrm{softmax}(A^u_t)
\end{align}
The model also can directly obtain useful information from the visual concept since visual concept features can provide object names/attributes in a textual semi-symbolic format.
\begin{align}
    (A^c_t)_{i} = e_t^{\top} C_i, \;\;\;
    a^c_t = \textrm{softmax}(A^c_t)
\end{align}

\vspace{3pt}
\par
\noindent\textbf{Selection Gate.}
To adaptively select the source of the next token, we employ gating approach~\cite{see2017get} to obtain the adaptive weights:
\begin{align}
    g_t &= \textrm{softmax}(W_g^{\top}e_t)
\end{align}
where $W_g \in \mathbb{R}^{d \times 3}$ is the trainable parameter. The weighted sum of each probability from each source is the final probability of the next token.
\begin{equation}
\begin{aligned}
    p(w_t|&w_{1:t-1}, I, D)= g_{t,0}\cdot a^g_t + g_{t,1}\cdot a^u_t+ g_{t,2}\cdot a^c_t
\end{aligned}
\end{equation}
The loss is: 
\begin{align}
    L &= -\sum_{t=1}^{T}\log{p(w_t^*|w_{0:t-1}, I, D)}
\end{align}
where $w_t^*$ is the GT token.

\section{Experiments}
\vspace{3pt}
\par
\noindent\textbf{Data Splits.}
We split the total 1,611 dialogues into 1,052, 262, and 297 for train, validation, and test set, respectively. From the dialogue splits, we obtain 4,059/1,002/1,112 (train/valid/test) instance splits.

\vspace{3pt}
\par
\noindent\textbf{Evaluation Metric.}
We use accuracy as the evaluation metric. For image search and editing systems, it is important to feed the correct command, and automatic metrics for text generation tasks such as BLEU~\cite{papineni2002bleu} are not appropriate. So, we only count generated commands which exactly match the ground-truth commands (i.e., command types and their arguments) as the correct ones. For the search command, generated commands with different query word orders (e.g., [\emph{search juice glass}] and [\emph{search glass juice}]) are also considered correct since queries with different word orders usually return the same or similar outcomes. For the color change command, we only compare the command type and color names but not up to intensity (e.g., [\emph{adjust\_color blue}]) since, in most cases, users ask to change colors without saying a specific value of intensity (e.g., \emph{``Color the image to the same color as the salmon in the above image''}). 

\begin{table}[t]
\begin{center}
\resizebox{0.9\columnwidth}{!}{
 \begin{tabular}{clc}
 \toprule
  & Models & Accuracy (\%)  \\
 \midrule
   1 & Request-Only & 42.30\\
    \midrule
    2 &DialogHistory-Only & 0.66  \\
    \midrule
   3 & Request+DialogHistory & 43.17 \\
    \midrule
  4 &  Vision-Only & 0.93 \\
    \midrule
   5 & Request+Vision & 45.56 \\
    \midrule
    6 & Request+DialogHistory+Vision & 46.43 \\
    \bottomrule
\end{tabular}
}
\end{center}
\caption{Modality ablations. Each modality/component helps improve the model's performance.  \label{tbl:unimodal}} 
\end{table}

\vspace{3pt}
\par
\noindent\textbf{Human Expert Performance.}
We randomly sample 50 instances for the search command and 10 instances for each of the other commands (total 100 instances) and ask an expert who knows the task well to predict the commands based on the textual and visual context.

\vspace{3pt}
\par
\noindent\textbf{Training Details.}
We use 512 as the hidden size and 256 as the word embedding dimension. We use Adam \cite{kingma2014adam} as the optimizer with the learning rate $1\times 10^{-4}$. See Appendix~\ref{app:training} for more details.

\begin{figure}[t]
    \centering
    \includegraphics[width=0.9\columnwidth]{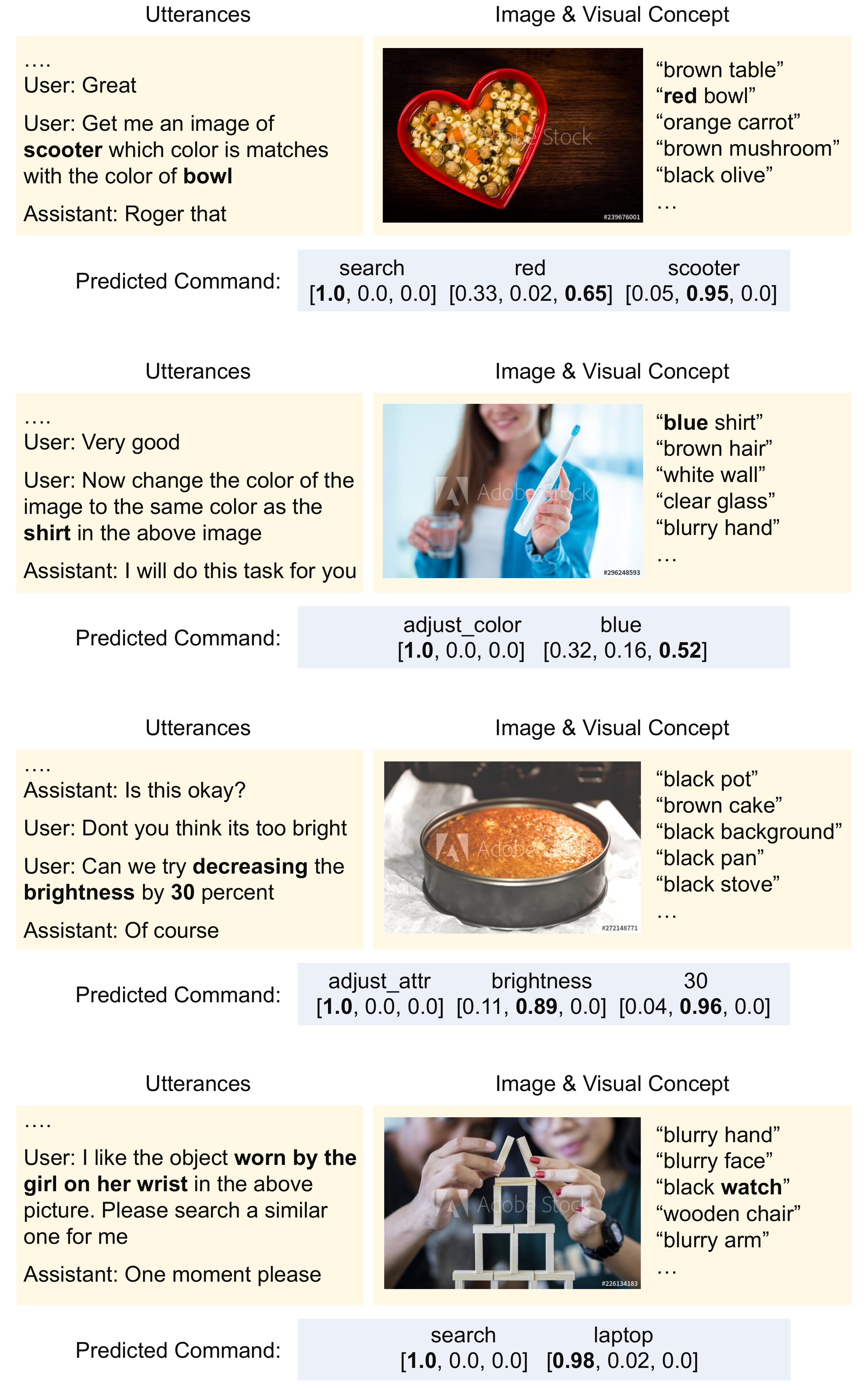}
    \caption{The examples of the model output (1st and 2nd examples: correct / 3rd and 4th: incorrect). Our model can effectively use the generator and extractors by selecting them with the adaptive selection gate (the numbers in bracket are the selection gate weight, i.e., [weight for the generator, weight for the utterance extractor, weight for the visual concept extractor]). The bottom two figures show incorrect examples in which the model cannot figure out the meaning of \emph{`decreasing'} and cannot catch \emph{`watch'} from the image. \label{fig:output_examples}}
\end{figure}

\section{Results}
As shown in Table \ref{tbl:result}, the extractor modules help improve the model's performance. The utterance extractor helps much to improve the model's performance (row 1 and 3). Especially, the scores for the search, brightness change, contrast change, and rotation commands get increased, implying that the utterance extractor can effectively locate the direct clues from the dialogue history context. Applying the visual concept extractor additionally increases the score (rows 3 and 4).\footnote{The stddev of the full model (Base+UE+VE) scores is 0.74, and the score of the model on validation split is 49.7\%.} The performance of the search and color change commands gets improved from this application, meaning that the visual concept extractor can match the visual features and the concept features, and align them with requests. But, when comparing rows 1 and 2, adding the visual concept extractor to the base model does not seem to help. Although it shows a similar improvement pattern for the other commands, the performance for the search command is not improved. That implies that the visual concept extractor is effective together with the utterance extractor (see the example at the top of Figure~\ref{fig:output_examples}).\footnote{While we evaluate the performance via the average score over each search/editing instance like in \cite{visdial}, one other possible evaluation option for practical applications is to consider the average success rate of the whole search/editing dialogue (5.4\% from our full (Base+UE+VE) model).}

\vspace{3pt}
\par
\noindent\textbf{Human Expert Performance.}
As shown in row 4 and 5 of Table~\ref{tbl:result}, the human-machine performance gaps are large for most of the command types, implying that there is large room for future work to develop novel improvements on this new multimodal dialogue task, and our baseline described above is meant to serve as a strong starting point. 

\vspace{3pt}
\par
\noindent\textbf{Modality Ablation.}
Table~\ref{tbl:unimodal} shows the ablation results from different combinations of the model (Base+UE+VE) components. We take the last two utterances from the dialogue as `request' since there is no explicit division between request and context in our \dataname{} dataset. As shown in row 2 and 4, the model could not perform well without the `request'. That is obvious since, without this information, the model cannot figure out what and how to search and edit. The request-only (row 1) records a high score possibly because many of the requests contain direct clues like \emph{``Can you rotate the image counterclockwise by 30 degrees''}. Adding dialogue history (row 1 and 3, row 5, and 6) helps, meaning the request needs dialogue context for better performance. Also, adding visual context (images) improves the model's performance (row 1 and 5, 3 and 6) because there are requests (such as for the search and color change commands) that need to refer to objects/colors in the visual context to be performed correctly.\footnote{We randomly sample 75 instances (except the first-turn search command) and conduct human evaluation on which inputs are required to perform the requests. Request-only: 43\%; need-DialogHistory+Vision: 57\% (need-DialogHistory 13\%, need-Vision 47\%, need-both 3\%). This means that to solve our command generation task, models need to understand the context (we observe the same trend when we also include the first search command).}

\vspace{3pt}
\par
\noindent\textbf{Output Examples.}
Figure~\ref{fig:output_examples} shows examples of the model output. In the top figure, our model gives the correct command ([\emph{search red scooter}]) according to the request. Specifically, the model generates the command name, \emph{`search'}, using the generator (with the selection gate weight of 1.0), extracts the color, \emph{`red'}, using the visual concept extractor (with the weight of 0.65), and also extracts the item name to search for, \emph{`scooter'} using the utterance extractor (with the weight of 0.95). The second figure shows the example of the color change command. The model also generates the correct command name, \emph{`adjust\_color'} using the generator (with the weight of 1.0). The model then extracts the color, \emph{`blue'}, from the visual concept feature using the visual concept extractor (with the weight of 0.52). On the other hand, as shown in the third figure, the model fails to understand the meaning of \emph{`decreasing'} and just extracts \emph{`30'} using the utterance extractor (with the weight of 0.96) for intensity (the ground-truth value is -30). In the bottom figure, the model cannot catch \emph{`watch'} in the image and generated the wrong searching query, \emph{`laptop'} using the generator (with the weight of 0.98). These negative results from our baseline model imply that there is room for improvement via more advanced modeling approaches in future work from the community on our \dataname{} task.

\section{Conclusion}
We introduced a novel conversational image search and editing task/dataset, called \dataname{}, in which an agent should conduct image search and editing according to users' requests. To implement and train the automated system, we collected a dialogue dataset in which a user and an assistant hold a conversation on image search/editing. We presented the generator-extractor model as a strong starting point baseline and the large human-machine performance gap showed there is room for improvement on this task.

\section*{Acknowledgments}
We thank the reviewers for their helpful comments. This work was partially done while HK was interning at Adobe Research and later extended at UNC, where it
was supported by NSF Award 1840131, ARO-YIP Award W911NF-18-1-0336, DARPA KAIROS Grant FA8750-19-2-1004, and a Google Focused Award. The views contained in this article are those of the authors and not of the funding agency. This work was done while DK was at Adobe Research. 
\bibliography{aaai22.bib}

\appendix
\section*{Appendices}
\begin{figure*}[t]
    \centering
    \includegraphics[width=1.90\columnwidth]{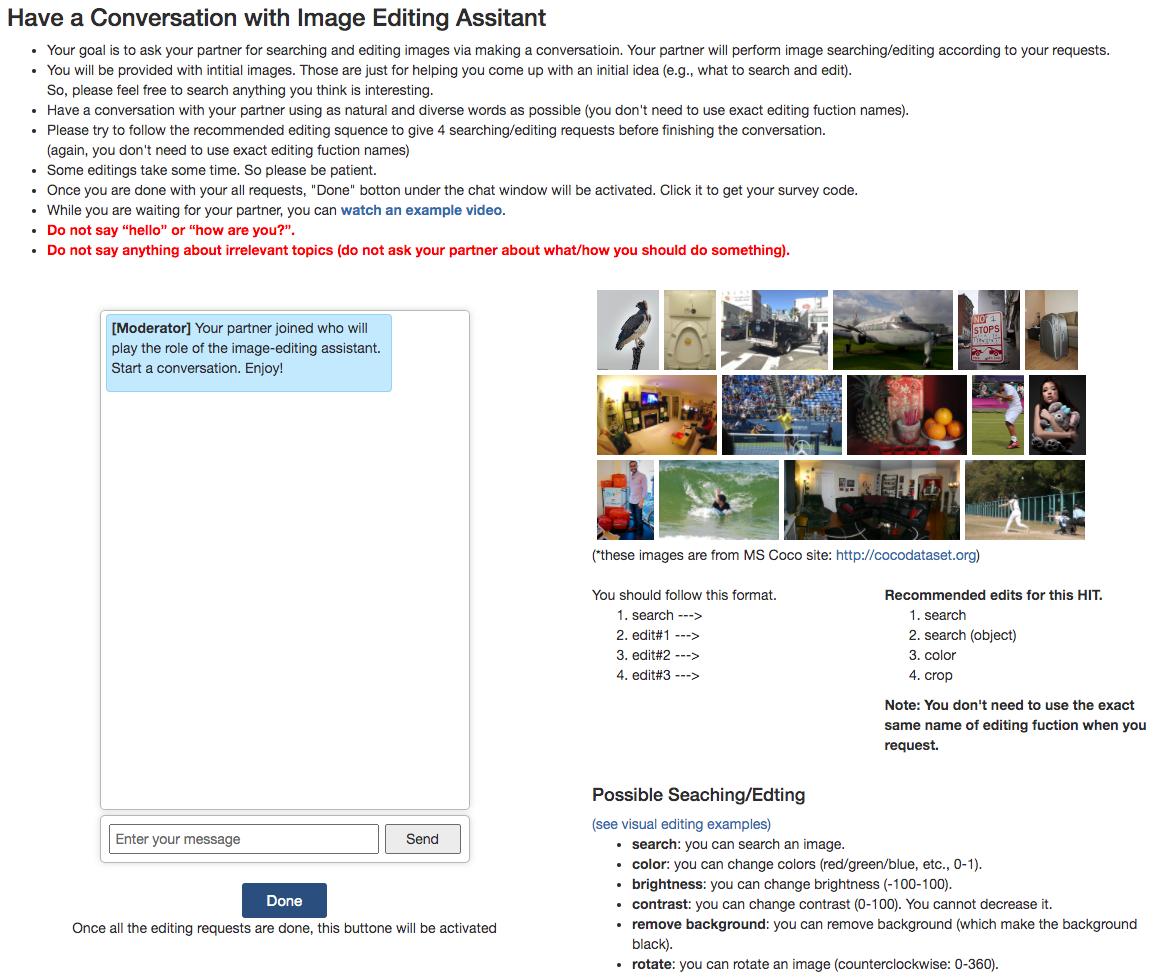}
    \caption{The data collection interface for user annotators.}
    \label{fig:user_interface}
\end{figure*}

\begin{figure*}[t]
    \centering
    \includegraphics[width=1.99\columnwidth]{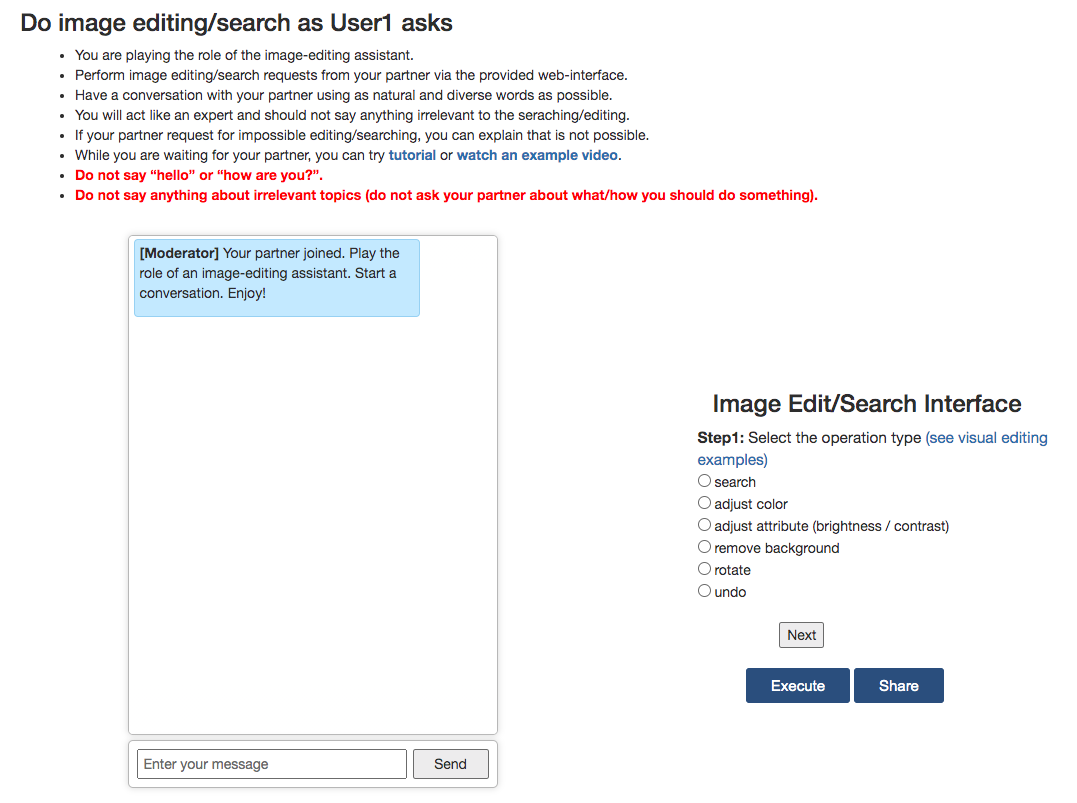}
    \caption{The data collection interface for assistant annotators.}
    \label{fig:assistant_interface}
\end{figure*}

\section{Data Collection Interface\label{app:interface}}
Figure~\ref{fig:user_interface} shows the data collection interface of user-annotators. They are provided with initial seed images and a list of suggestions for image search/editing functions. Figure~\ref{fig:assistant_interface} shows the assistant-annotators' interface. They are provided with tools for image search and editing functions.

\section{Annotation Quality Control\label{app:annotator}}
To ensure the annotations maintain high quality, we train and monitor the annotations through an intensive training session. Specifically, one of the authors of this paper is directly paired with some representative annotators (we only know the names of the representative annotators and do not know any private information of them and the other annotators) and collects dozens of dialogues for practice and trains them by correcting every single error. Then the representative annotators train other annotators and collect practice dialogues from each pair of the annotators. We check all the practice dialogues manually and give feedback. We perform this training session multiple times until the quality of the dialogues gets above some threshold (i.e., until the dialogues have no obvious/critical issue).

\section{Training Details (Reproducibility)\label{app:training}}
All the experiments are run on a Ubuntu 16.04 system using the NVIDIA GeForce GTX 1080 Ti GPU and Intel Xeon CPU E5-2630. We use PyTorch \cite{paszke2017automatic} to build models. We use 512 as the hidden size, 256 as the word embedding dimension, and 2048 as the visual feature (Faster R-CNN) dimension. We use Adam \cite{kingma2014adam} as the optimizer with the learning rate $1\times 10^{-4}$. For dropout p values, 0.3 and 0.5 are used (each for different layers). We run 500 epochs (each epoch takes around 1min 10secs including training+evaluation) for each experiment for the model selection. We use 2020/2021/2022 as random seed values. All the scores are the average values from the run with the three different seeds. The number of trainable parameters of our full model is 11.4M. We use manual tuning (e.g, learning-rate=\{$1\times 10^{-4}$, ..., $1\times 10^{-6}$\}, dropout=\{0.3. 0.5\}, etc.) for choosing hyperparameter values (based on validation scores).

\end{document}